\documentclass[lettersize,journal]{IEEEtran}
\usepackage{amsmath,amsfonts}
\usepackage{algorithm}
\usepackage{algpseudocode}
\usepackage{array}
\usepackage[caption=false,font=normalsize,labelfont=sf,textfont=sf]{subfig}
\usepackage{textcomp}
\usepackage{stfloats}
\usepackage{url}
\usepackage{verbatim}
\usepackage{graphicx}
\usepackage{cite}
\usepackage{booktabs}

\usepackage{multirow}
\usepackage{multicol}

\usepackage{colortbl}
\usepackage[table]{xcolor}
\usepackage{makecell}

\begin{document}

\title{From Competition to Coopetition: Coopetitive Training-Free Image Editing Based on Text Guidance}

\author{
Jinhao Shen,~Haoqian Du,~Xulu Zhang, Xiao-Yong Wei,~\IEEEmembership{Senior Member, IEEE} and Qing Li,~\IEEEmembership{Fellow, IEEE}

\thanks{Jinhao Shen, Haoqian Du, Xulu Zhang and Qing Li are with the Department of Computing, Hong Kong Polytechnic University, Hong Kong, China (e-mail: jinhao00.shen@connect.polyu.hk; duhaoqian879@gamil.com; compxulu.zhang@connect.polyu.hk; csqli@comp.polyu.edu.hk)}
\thanks{Xiao-Yong Wei is with the Department of Computer Science, Sichuan University, Chengdu 610017, China, and also with the Department of Computing, Hong Kong Polytechnic University, Hong Kong, China (e-mail: cswei@scu.edu.cn). \textit{(Corresponding author: Xiao-Yong Wei.)}}
}

\maketitle

\begin{abstract}

Text-guided image editing, a pivotal task in modern multimedia content creation, has seen remarkable progress with training-free methods that eliminate the need for additional optimization. Despite recent progress, existing methods are typically constrained by a competitive paradigm in which the editing and reconstruction branches are independently driven by their respective objectives to maximize alignment with target and source prompts. 
The adversarial strategy causes semantic conflicts and unpredictable outcomes due to the lack of coordination between branches. 
To overcome these issues, we propose Coopetitive Training-Free Image Editing (CoEdit), a novel zero-shot framework that transforms attention control from competition to coopetitive negotiation, achieving editing harmony across spatial and temporal dimensions. 
Spatially, CoEdit introduces Dual-Entropy Attention Manipulation, which quantifies directional entropic interactions between branches to reformulate attention control as a harmony-maximization problem, eventually improving the localization of editable and preservable regions. 
Temporally, we present Entropic Latent Refinement mechanism to dynamically adjust latent representations over time, minimizing accumulated editing errors and ensuring consistent semantic transitions throughout the denoising trajectory. 
Additionally, we propose the Fidelity-Constrained Editing Score, a composite metric that jointly evaluates semantic editing  and background fidelity. Extensive experiments on standard benchmarks demonstrate that CoEdit achieves superior performance in both editing quality and structural preservation, enhancing multimedia information utilization by enabling more effective interaction between visual and textual modalities. The code will be available at \url{https://github.com/JinhaoShen/CoEdit}.

\end{abstract}


\section{Introduction}

Recent advancements in generative models have unlocked a wide array of text-conditioned vision tasks, including high-fidelity image generation \cite{imggen1, zhang2024compositional}, restoration \cite{restoration1, restoration2}, understanding \cite{WXY_CAL_TIP_2013, jiang2024prior, xiaoyongwei2012acmmm} and editing~\cite{10175586, 10480591, 10964679}. 
Among these, text-conditioned image editing has emerged as a critical capability, enabling users to intuitively modify image attributes, add or remove objects, or change styles via natural language descriptions while preserving the underlying image structure. 
While conventional approaches often demand extensive, task-specific training on large-scale datasets, the advent of training-free methods~\cite{trainfree1, trainfree2} presents a compelling paradigm shift. 
By obviating the need for model retraining or fine-tuning, these methods significantly enhance operational efficiency and accessibility, offering a powerful alternative for flexible image editing.

The main training-free image editing methods predominantly employ Stable Diffusion \cite{song2020denoising}, utilizing attention-based control mechanisms. 
Prompt-to-Prompt \cite{ptop}, PnP \cite{pnp} demonstrate that some of attention maps can capture the relationship between tokens and image regions.

The training-free image editing pipeline is divided into two branches: a reconstruction branch that recovers the original image conditioned on its source prompt, and an editing branch where target prompt attention maps replace the original ones in denoising network. 

\begin{figure}[t]
    \centering
    \includegraphics[width=1.0\linewidth]{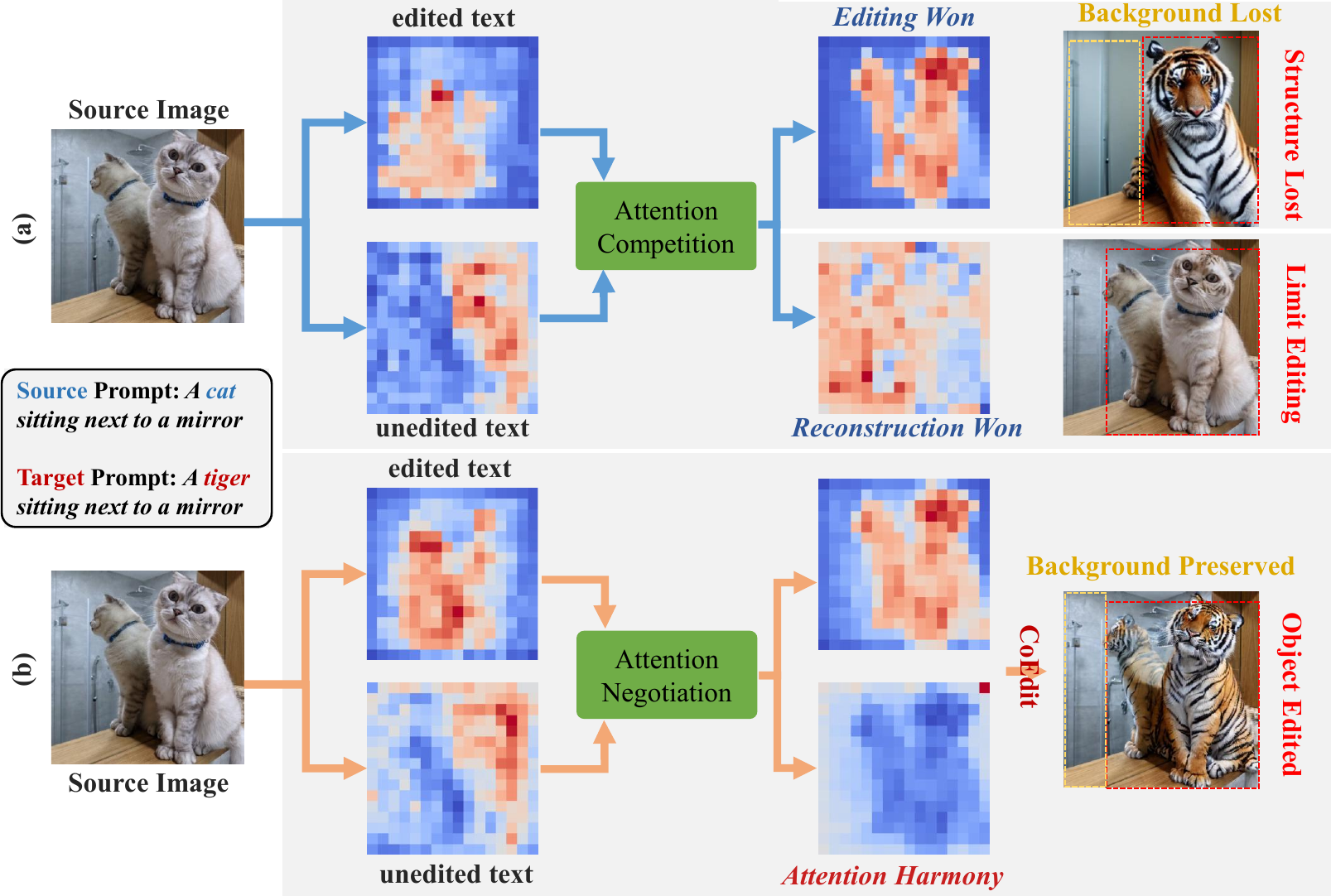}
    \caption{Difference between competitive and coopetitive strategies. (a) Competitive strategy causes two typical failure modes: background degradation when the editing branch dominates, and insufficient editing when the reconstruction branch prevails. (b) CoEdit adopts the coopetitive strategy that enables attention negotiation between the branches.}
    \label{fig:motivation}
\end{figure}

\begin{figure}[t]
    \centering
    \includegraphics[width=0.95\linewidth]{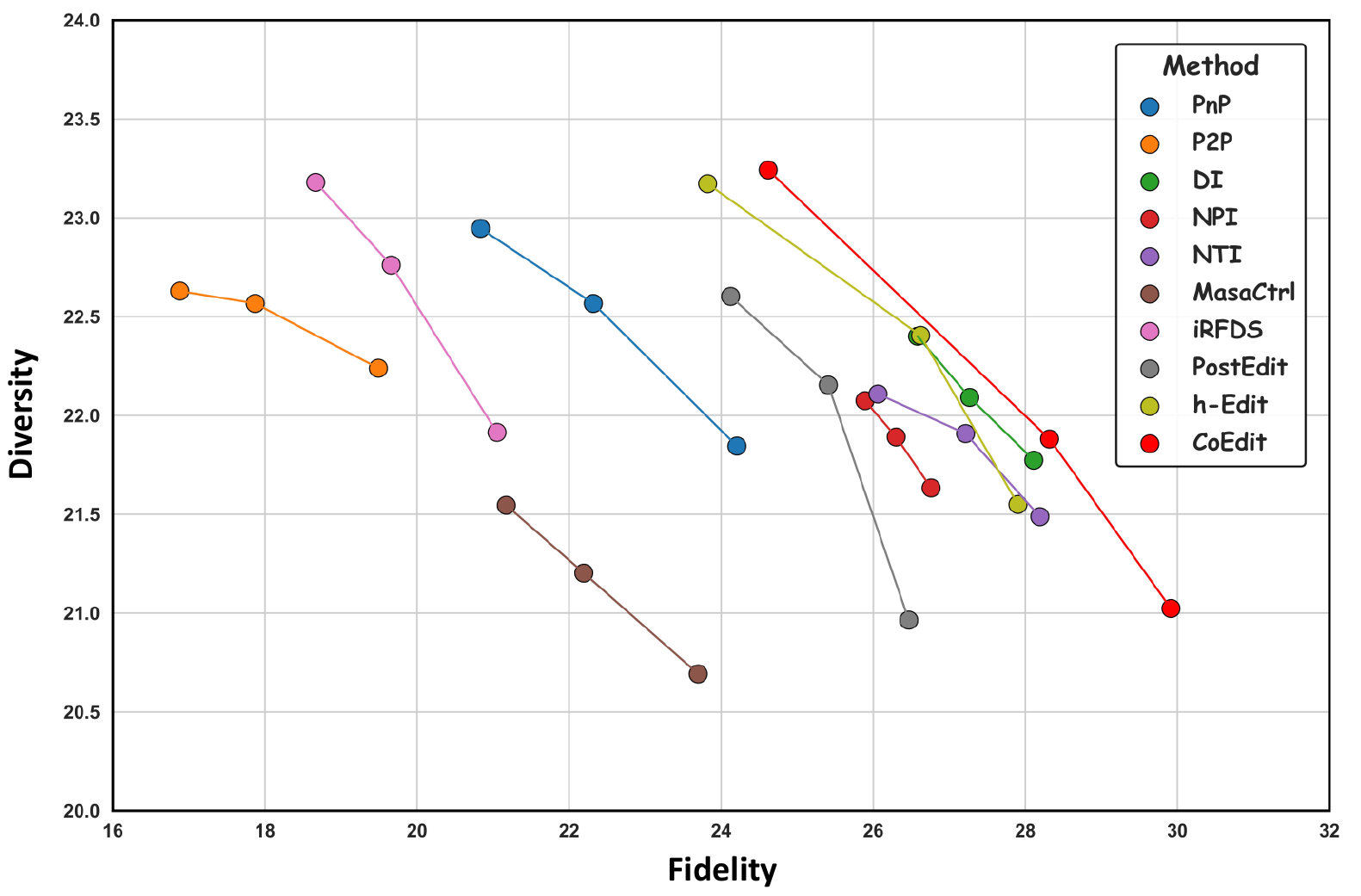}
    \caption{
    Quantitative trade-off between editing diversity ($\text{CS}_r$) and reconstruction fidelity (PSNR) under default, editing, and reconstruction settings. 
    CoEdit is located in the top-right region of the plot, indicating its capability to achieve both superior editing accuracy and high reconstruction fidelity.}
    \label{fig:line_performance}
\end{figure}

However, existing attention-based editing frameworks are inherently constrained by their reliance on a \textbf{competitive strategy} in which the editing and reconstruction branches independently and greedily pursue the maximization of their respective alignment with the target and source prompts, without mutual coordination.
For instance, Fig.~\ref{fig:motivation}(a) illustrates the inherent limitations of purely competitive strategies in attention-based image editing. 
The source image is  processed through the iterative denoising steps, and we visualize attention scores associated with both edited and unedited textual tokens.
After attention competition, the edited image often demonstrates distinct failure modes \cite{hedit, Gomez_Trenado_2025}: (1) \textit{editing branch won}: background regions are undesirably altered and the target object (e.g., a tiger) loses structural integrity \cite{iRFDS, PostEdit}; (2) \textit{reconstruction branch won}, the edited image retains features of the original subject (e.g., a cat), failing to reflect the intended semantic modification \cite{xu2023infedit}.
These issues motivate an essential question:
\textit{Can we move beyond competition toward a coopetitive mechanism, where editing and reconstruction cooperate for mutual benefit?}

To address this, we propose \textbf{Coopetitive Training-Free Image Editing (CoEdit)}, a novel zero-shot image editing framework that transforms attention control from a competitive paradigm into a \textbf{coopetitive strategy}, as illustrated in Fig.~\ref{fig:motivation}(b). 
CoEdit encourages the editing and reconstruction branches to engage in structured competition over attention allocation while jointly negotiating semantic influence through entropy-guided cooperation across spatial and temporal dimensions, as detailed below. 

From the spatial perspective, although coopetition is conceptually appealing, it lacks a principled formulation to quantify each branch’s editing demand and competitive intensity. 
The interaction is modeled as a pixel-level competition at each step, where both branches seek to claim semantic ownership over overlapping regions, leading to inconsistent attention focus and semantic interference between editable and preservable areas. Hence, we introduce Dual-Entropy Attention Manipulation, which measures the directional entropy extracted from both branches. It quantifies the coherence of their attention assignments and reformulates the objective as a harmony-maximization problem over editable and preservable regions.

While spatial negotiation facilitates localized editing balance at individual time steps, it is insufficient alone for ensuring long-range consistency.
In the early stages of denoising, rigid and erroneous editing boundaries often cause unpredictable interactions in spatial negotiation, which in turn lead to structural drift and textural collapse in editable regions as generation progresses. 

To mitigate these cumulative distortions, CoEdit introduces Entropic Latent Refinement, which smooths semantic transitions over time by dynamically adjusting latent representations. 
The temporal coordination minimizes editing errors across the entire denoising trajectory, ensuring coherent structural evolution and consistency in the editing trajectory.

Comprehensive experiments demonstrate that CoEdit strikes a superior balance between editing accuracy and reconstruction fidelity, positioning it in the top-right of the performance trade-off plot (Fig.~\ref{fig:line_performance}). 

Recognizing that existing metrics inadequately capture this crucial trade-off, we further propose the Fidelity-Constrained Editing Score (FCES). 
The novel, F1-inspired metric holistically evaluates performance by concurrently assessing the quality of semantic edits and the fidelity of structural preservation.

In summary, the main contributions of this paper are as follows:
\begin{itemize}
    \item We propose CoEdit, a novel training-free image editing framework that pioneers a coopetitive strategy, fundamentally resolving the destructive interference between editing and reconstruction branches found in prior attention-based methods.
    \item We introduce two synergistic mechanisms: Dual-Entropy Attention Manipulation for principled spatial negotiation and Entropic Latent Refinement for ensuring global consistency across the denoising process.
    \item We design and present a new evaluation metric, the Fidelity-Constrained Editing Score (FCES), which provides a more balanced and comprehensive assessment of the trade-off between editing accuracy and reconstruction fidelity.
\end{itemize}

\section{Related work}
\subsection{Attention-Based Text-Guided Image Editing}  
A series of methods achieve controllable image editing by manipulating attention mechanisms within diffusion models.  
These approaches typically operate on cross-attention~\cite{ptop,customedit,foi, zhang2025revisiting} or self-attention~\cite{pnp,masactrl} to guide the model toward desired edits while preserving structural consistency. For example, Prompt-to-Prompt (P2P)~\cite{ptop} injects cross-attention maps from a reference trajectory, and Plug-and-Play (PnP)~\cite{pnp} performs similar manipulation on self-attention to maintain spatial layout.  
Recent advances further explore mutual attention~\cite{masactrl,tmmattention1,tmmattention2, attention1pami, tmmattention3,attention2tip}, inversion-enhanced control~\cite{duan2024tuningfree,pan2023effective,huberman-spiegelglas2024editfriendly} and instant attention masks~\cite{zou2024towards, attention3, nguyen2025swiftedit} to improve edit fidelity, convergence speed, and control strength.  

In particular, Duan et al.\ propose Tuning‑Free Inversion‑Enhanced Control ~\cite{duan2024tuningfree}, which extracts key and value features from self-attention during DDIM inversion and injects them into the sampling process—thus achieving consistent real‑image editing without any fine‑tuning of the diffusion model.  
Qiao et al.\ propose BARET~\cite{qiao2023baret}, introducing a Balanced Attention Module (BAM) that fuses the self-attention maps from the reconstruction branch with the cross-attention maps from the transition (editing) branch, thereby balancing textual description and image semantics to optimize target‑text guidance.  
Zou and Tang et al.\ propose Instant Attention Masks~\cite{zou2024towards}, dynamically generating attention masks on-the-fly to accelerate diffusion-based semantic image editing without sacrificing spatial precision.  

Specifically, the self- and cross-attention operations are defined as:
\begin{equation} \label{eq:cross_attn}
  \{Q_{t, i}^{s}, K_{t, i}^{s}, V_{t, i}^{e}\}
  = \text{CA}(Q_{t, i}^e, K_{t, i}^e, Q_{t, i}^s, K_{t, i}^s),\; V_{t, i}^e,
\end{equation}
\begin{equation} \label{eq:self_attn}
  \{Q_{t, i}^e, K_{t, i}^s, V_{t, i}^{s}\}
  = \text{SA}(Q_{t, i}^e, K_{t, i}^e, Q_{t, i}^s, K_{t, i}^s),\; V_{t, i}^s,
\end{equation}
where the self-attention step $\text{SA}(*)$ transfers structural priors from the reconstruction branch, and the cross-attention step $\text{CA}(*)$ injects target semantics from the editing branch. It is evident that previous attention-based editing methods replace the maps of the reconstruction branch with those of the editing branch. Consequently, they neither effectively mitigate inter-branch contention nor achieve sufficiently fine-grained spatial partitioning of image regions.

\subsection{Consistency Sampling of Denoising Diffusion}
Consistency models \cite{song2023consistency, song2023improved, wang2024animatelcm,starodubcev2024invertible,liu2024scott} facilitate the emergence of a series of novel generative models by effectively distilling knowledge from pre-trained Stable Diffusion.
They enforce the self-consistency property \cite{song2023consistency}, which ensures that any point along the same probability flow ordinary differential equation trajectory maps to the same solution. 
It has been extended to enable high-resolution text-to-image synthesis through the introduction of latent consistency models \cite{luo2023latent}. 
PostEdit~\cite{PostEdit} adapts consistency models by configuring DDIM with \( \sigma_t = \sqrt{1 - \alpha_{t-1}} \), under which the update rule can be reformulated as:
\begin{equation} \label{eq:ddim_2}
    z_{t-1} = \sqrt{\frac{\alpha_{t-1}}{\alpha_{t}}}\left(z_{t}-\sqrt{1-\alpha_{t}}\varepsilon_{\theta}(z_{t},t)\right) + \sqrt{1-\alpha_{t-1}}\epsilon_t,
\end{equation}
where $\epsilon_t\sim\mathcal{N}(0,\mathbf{I})$, and we assumes the first term serves as an approximation of $z_{0}$.
Following existing work \cite{xu2023infedit}, \( z^s_{t-1} \) from the reconstruction branch can be directly derived by substituting \( z^s_t \) and \( \epsilon^s_t \) into Eq.~\ref{eq:ddim_2}.
The reverse process of $z^e_t$ is defined as:
\begin{align} \label{eq:ddcm_sample}
z^e_{t-1} &= \sqrt{\alpha_{t-1}}z_0 + \sqrt{\alpha_{t-1}/\alpha_{t}}(z^e_t-z^s_t) \\ \nonumber
& \quad + \sqrt{(1-\alpha_t)\alpha_{t-1}/\alpha_t}(\epsilon^e_t-\epsilon^s_t) \\ \nonumber
& \quad + \sqrt{1-\alpha_{t-1}}\cdot\epsilon_t.
\end{align}

While this formulation improves sampling consistency, it overly prioritizes alignment with the original image latent.
Instead of reconstructing the image as a whole, we emphasize the importance of spatially partitioned processing.

\subsection{Editing Latent Refinement} 
A series of studies~\cite{mcallister2024rethinking, dds, sauer2024adversarial,islamfahim2025nulltoon,hwang2025qsd,NEURIPS2024_da9e48b4} propose refinement objectives to guide the evolution of the editing image latent, all while meticulously ensuring that the parameters of the underlying model are completely frozen.
In this work, the general formulation of editing latent refinement can be expressed as:
\begin{equation} \label{eq:optim_edit}
\mathbf{z}_{t-1} \leftarrow \left[ \mathbf{z}_{t} - s_{t} \nabla_{\mathbf{z}_{t}} \mathcal{L}_{\text{opti}} \right]_{i=1}^{N_{0}},
\end{equation}
where \( \mathcal{L}_{\text{opti}} \) denotes the editing objective, \( s_t \) is the gradient scale, and \( N_0 \) indicates the number of refinement steps at timestep \( t \). This framework updates the latent code \( \mathbf{z}_t \) through iterative gradient descent, guided by user-defined objectives to enforce desired modifications.
Notably, recent methods \cite{dds,PostEdit, hedit} primarily optimize the entire latent representation to compute $\mathcal{L}_{\text{opti}}$.
PostEdit~\cite{PostEdit} leverages a posterior sampling scheme within an optimization framework to efficiently guide the diffusion sampling process for zero-shot image editing.

The editing latent loss in these schemes is often subject to considerable interference from background noise. 
Following Eq.~\ref{eq:optim_edit}, we introduce structural and textural semantic corrections from a coopetitive perspective.

\section{Methodology}
\begin{figure*}
    \centering
    \includegraphics[width=0.9\linewidth]{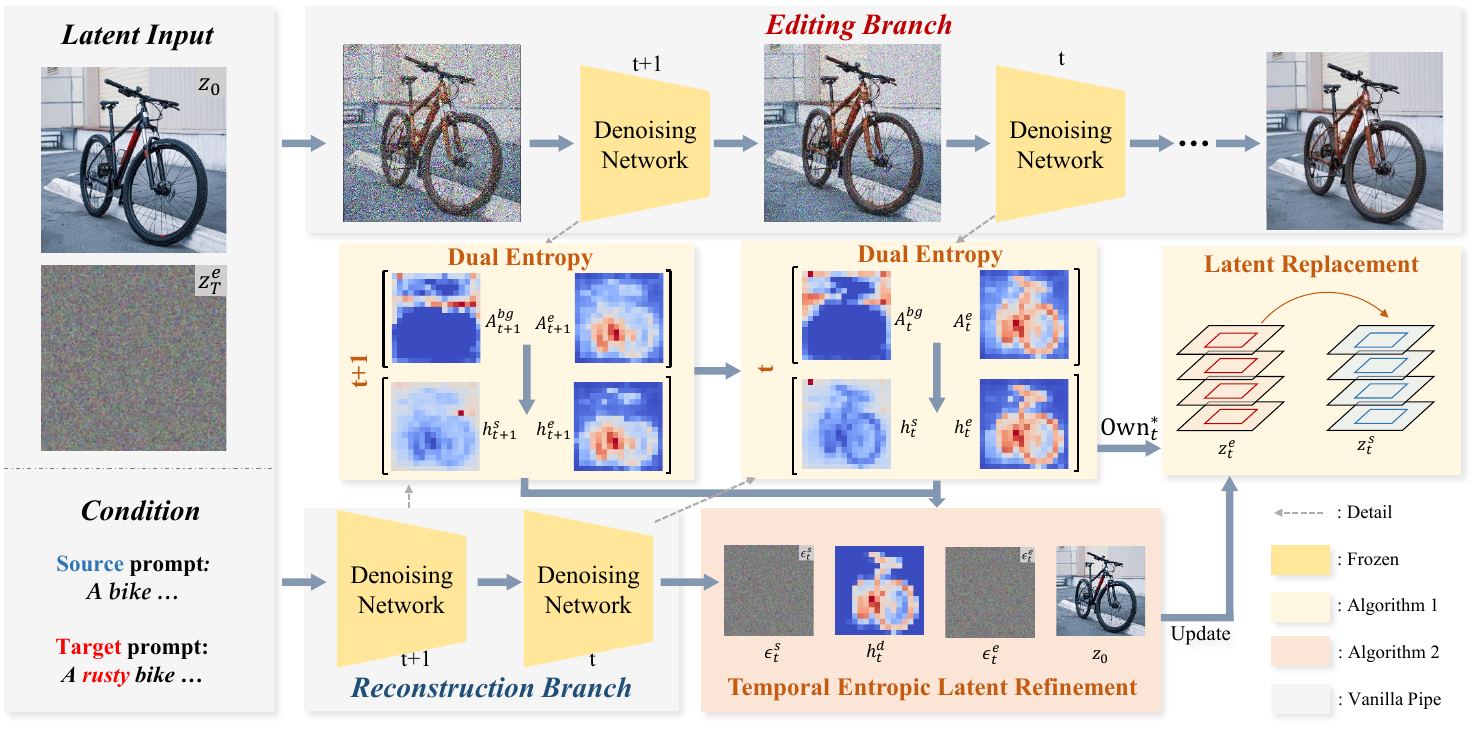}
    \caption{Overview of the proposed CoEdit framework, which integrates Dual-Entropy Attention Manipulation and Entropic Latent Refinement.}
    \label{fig:Overview}
\end{figure*}

\subsection{Preliminaries}
In existing training-free editing pipelines, semantic competition predominantly emerges between the editing and reconstruction branches during denoising. Formally, the source image latent \( z_0 \) serves as the input to both branches: the editing branch is conditioned on the target prompt \( \mathcal{C}^e \), while the reconstruction branch is guided by the source prompt \( \mathcal{C}^s \).

At each sampling timestep \( t \in \mathcal{T}=[1, 2,..., t, ..., T]\), this dual-branch structure independently predicts attention maps and noise components:
\begin{equation} \label{eq:denoise}
    \epsilon^{s}_t, A^{s}_t = \varepsilon_\theta(z^{s}_t, \mathcal{C}^s, t), \quad 
    \epsilon^{e}_t, A^{e}_t = \varepsilon_\theta(z^{e}_t, \mathcal{C}^e, t),
\end{equation}
where \( A^{s}_t \) and \( A^{e}_t \) denote the normalized attention maps derived from the reconstruction and editing branches, respectively. 
We define the ownership \( M_t \) based on $A^e_t$ to assign editing and reconstruction regions:
\begin{equation} \label{eq:spatial_owner}
M_t = \text{Own}_t({A^e_t}),
\end{equation}
\begin{equation} \label{eq:z_replace}
z^e_t = M_{t} \cdot z^e_t + (1 - M_{t}) \cdot z^s_t.
\end{equation}
Here, \( \text{Own}_t(*) \) denotes a coopetition function that operates on \( A^e_t \) to assign editable regions at timestep $t$, with $M_t$ encoding the coopetitive outcome between two branches. 
Existing works \cite{PostEdit, xu2023infedit} adopt a hard binary masking strategy, wherein pixels with attention values exceeding a fixed threshold are designated as editable.

Temporal competition emerges during the sampling transition from \( t \) to \( t\!-\!1 \) in Eq.\ref{eq:ddcm_sample} as the stochastic noise introduces semantic ambiguity and disrupts editing consistency, which ultimately hinders the delineation of regional ownerships.

Building upon the aforementioned competition strategies, we define a unified coopetition objective as follows: 
\begin{equation} \label{eq:coop_objective}
Obj_{coop} = \underset{\{h_{}^{e}, h_{}^{s}\}}{\min} \sum_{t\in \mathcal{T}} \{ | h^e_t \cdot h^s_t |+ \\
| \text{Own}_t(A^e_t) - M_{T} | \} ,
\end{equation}

where $\mathcal{T}$ denotes the set of all sampling timesteps. 
\( h^e_t \cdot h^s_t \) denotes the element-wise product of the entropy maps, which encourages \( h^e_t \) and \( h^s_t \) to focus respectively on the editing and reconstruction objectives. \( M_{T} \) denotes the ownership mask obtained at the final denoising step.
Based on the coopetition objective, we construct the CoEdit framework, as illustrated in Fig.~\ref{fig:Overview}.

\subsection{Spatial Coopetition Enforcement} 
Spatial competition naturally arises from the adversarial interactions between the editing and reconstruction branches, which contend for control over editable and preservable regions.
CoEdit begins with the quantification of spatial competition with \( h^{e}_t \) and \( h^{s}_t \), to capture the distinct semantic intents of the editing and reconstruction branches, as formulated in Algorithm~\ref{alg:spatial}.
We first extract the normalized cross-attention maps before Eq.\ref{eq:cross_attn}, referred to as $A^{s}_{t}$ and $A^{e}_{t}$.
$A^{s}_{t}$ represents the preserved semantics of unmodified texts, while $A^{e}_{t}$ denotes the editing content of modified lexical items.
Since the editing information is coupled with $A^{s}_{t}$, we apply directional differencing to derive a clean background attention map, denoted as $A^{bg}_{t}$.
\begin{equation} \label{eq:A_bg}
    A^{bg}_t=\text{ReLU}(A^{s}_t - A^{e}_t).
\end{equation}
Information entropy naturally quantifies uncertainty and serves as a foundation for regulating spatial interactions under a coopetitive framework.
Hence, we define the editing-direction entropy, a directional component of cross-entropy, to guide the editing dynamics.
It is formulated as follows:
\begin{equation} \label{eq:entropy_edit}
    h^{e}_{t} = -A^{e}_{t}\text{log}(A^{bg}_{t}),
\end{equation}
where $h^{e}_{t}$ is a direction-aware variant of cross-entropy to regulate the editing dynamics during sampling, namely editing-direction entropy. 
Rather than competitive strategy, which treat editing and preservation as mutually exclusive, we view them as complementary forces, namely spatial coopetition.
The preservation of structural background regions reinforces the editing process by promoting spatial harmony.
Furthermore, we also define the reconstruction-direction entropy as follows:
\begin{equation} \label{eq:entropy_reconstruct}
    h^{s}_{t} = -(1-A^{e}_{t})\text{log}(1-A^{bg}_{t}).
\end{equation}

\begin{algorithm}[t]
\caption{Spatial Coopetition via Dual-entropy Attention Manipulation}
\label{alg:spatial}
\begin{algorithmic}[1]
\State \textbf{Input:}
\State \quad Denoising Network: $\varepsilon_\theta$ 
\State \quad Timestep: $t$ \ ; \quad Source Input Latent: $z_0$
\State \quad Source\ /\ Target Prompts: $\mathcal{C}^s/\ \mathcal{C}^e$
\State \quad Latent of Reconstruction\ /\ Editing Branchs: $z^s_t/\ z^e_t$
\For{$t=1$ \textbf{to} $T$}
\State $\epsilon^{e}_t,A^{e}_t=\varepsilon_\theta(z^{e}_t,\mathcal{C}^e, t)$
\State $\epsilon^{s}_t,A^{s}_t=\varepsilon_\theta(z^{s}_t,\mathcal{C}^s,t)$
\State \texttt{/* Dual Entropy */}
\State $ h^{e}_{t}, h^{s}_{t}= -A^{e}_t\text{log}(A^{bg}_t), -(1-A^{e}_t)\text{log}(1-A^{bg}_t)$
\State \texttt{/* Spatial Coopetition*/}
\State  $ M_t^* = \text{Own}^*_t(\frac{\sum_{i=1}^{p} \left( {S_i}/{S_p} \right) - 1}{p} \times \frac{\ell(h^{e}_{t})}{\ell(h^{e}_{t})+\ell(h^{s}_{t})}+$
\State  $ \frac{1}{2}(\ell(h^{e}_{t})+\ell(h^{s}_{t}))) $
\State \texttt{/* Latent Replacement */}
\State $z^e_t = M^*_{t}\cdot z^e_t + (1-M^*_{t})\cdot z^s_t$
\EndFor
\end{algorithmic}
\end{algorithm}

Dual-entropy formulations enable precise quantification of spatial coopetition, guiding the interplay between image editing and structural preservation in a complementary manner.
We design an entropy-guided threshold that incorporates the structural bias and adaptive modulation, and subsequently define the ownership estimator \( G \), which constitutes the second component of the objective in Eq.~\ref{eq:coop_objective}.
In detail, we first reshape the attention map $A^e_t$ into a vector of length $p = H \times W$ and sort it in ascending order, denoted as ${s_1 \leq s_2 \leq \dots \leq s_p}$.
The normalized cumulative sum is then computed as $S_i = \sum_{j=1}^{i} s_j$, with $S_p$ denoting the total sum.
Subsequently, the spatial-aware ownership estimator $G_{t}$ is estimated from the cumulative distribution as:
\begin{align} \label{eq:g_thres}
\text{Own}^*_t({A^e_t}) ={} & p^{-1} \left( \sum_{i=1}^{p} \left( {S_i}/{S_p} \right) - 1 \right) \frac{\ell(h^{e}_{t})}{\ell(h^{e}_{t}) + \ell(h^{s}_{t})} \nonumber \\
      & + \frac{1}{2}(\ell(h^{e}_{t}) + \ell(h^{s}_{t})). 
\end{align}
Here, $\ell(*)$ represents the coopetition norm function. 
$\frac{1}{2}(\ell(h^{e}_{t}) + \ell(h^{s}_{t}))$ reflect the bias of entropic signals.
Unlike existing approaches, our designed $\text{Own}^*_t(*)$ provides a dynamic thresholding mechanism to generate binary ownership mask \( M^*_t\).
We further leverages \( M^*_t\) to fuse latent features, such that the negotiated spatial semantics are reliably preserved by applying \( M_t^* \) within the update equation given in Eq.~\ref{eq:z_replace}.

\subsection{Temporal Coopetition Coordination}
While spatial cooperation facilitates localized and structurally consistent modifications, temporal coordination remains critical for maintaining semantic continuity and guiding dynamic adaptation throughout the denoising trajectory.
Thus, we extend our entropy-based framework from spatial to temporal dimensions, as detailed in Algorithm~\ref{alg:opti}. 
It aligns with Eq.~\ref{eq:coop_objective} to minimize temporal inconsistency over \( t \in \mathcal{T} \).

The normalized $h^{s}_{t}$ and $h^{e}_{t}$ encode multiscale spatial semantics, with their temporal variations capturing the evolving structural and textural dynamics of the edited and preserved regions.
These variations are leveraged to dynamically adjust latent representations, thereby ensuring temporal consistency and capturing structural deviation across the denoising trajectory.
Therefore, we quantify the temporal entropy divergence as follows: 
\begin{gather} \label{eq:entropy_time}
    h^{ed}_{t},\ h^{sd}_{t} = h^{e}_{t} - h^{e}_{t+1},\ h^{s}_{t} - h^{s}_{t+1}, \\
    h^{d}_{t} = h^{ed}_{t} - h^{sd}_{t}.
\end{gather}
Here, $h^{ed}_{t}$ and $h^{sd}_{t}$ represent the temporal variations of directional entropy within the editing and background regions, respectively. 

The cross-step entropy divergence $h^{d}_{t}$ characterizes semantic misalignment between the editing and reconstruction regions over time, serving as a coordination signal for regulating structural evolution and suppressing cumulative inconsistency.

\begin{algorithm}[t]
\caption{Temporal Coopetition via Entropic Latent Refinement}
\label{alg:opti}
\begin{algorithmic}[1]
\State $h^{e}_{t},\ h^{s}_{t} = -A^{e}_t \log A^{bg}_t,\ -(1 - A^{e}_{t}) \log (1 - A^{bg}_t)$
\State $h^{e}_{t+1},\ h^{s}_{t+1} = -A^{e}_{t+1}\!\log\!A^{bg}_{t+1},\ -(1\!-\!A^{e}_{t+1})\!\log(1\!-\!A^{bg}_{t+1})$

\State \texttt{/* Directional Entropy */}
\State $h^{ed}_{t},\ h^{sd}_{t} = h^{e}_{t} - h^{e}_{t+1},\ h^{s}_{t} - h^{s}_{t+1}$
\State \texttt{/* Cross-step Entropy Divergence */}
\State $h^{d}_{t} = h^{ed}_{t} - h^{sd}_{t}$
\For{$i = 1$ \textbf{to} $N_o$}
    \State $\epsilon^{e}_t,\ A^{e}_t = \varepsilon_\theta(z^{e}_t,\ \mathcal{C}^e,\ t)$
    \State $A^{bg}_t = \text{ReLU}(A^{s}_t - A^{e}_t)$
    \State $h^{e}_{t},\ h^{s}_{t} = -A^{e}_t \log A^{bg}_t,\ -(1 - A^{e}_t) \log (1 - A^{bg}_t)$
    \State \texttt{/* Temporal Coopetition Noise */}
    \State $\epsilon^{dir}_{t} = \sqrt{\frac{1 - \alpha_t}{\alpha_t}} (\epsilon^e_t - \epsilon^s_t + \frac{h^{d}_{t} - \text{mean}(h^{d}_{t})}{\text{std}(h^{d}_{t})})$
    \State \texttt{/* Editing Latent Refinement */}
    \State $z_t^e = z_t^e - \nabla_{z_t^e}(\epsilon^{dir}_{t} \cdot z^e_t \cdot M^*) + h^{d}_{t}$
\EndFor
\State Sample noise $\epsilon \sim \mathcal{N}(0,I)$
\State $z_{t-1} = \sqrt{\alpha_{t-1}} z_0 + \sqrt{\alpha_{t-1}} (z^e_t - z^s_t) M^*_t + \sqrt{(1 - \alpha_t){\alpha_{t-1}}/{\alpha_t}} (\epsilon^e_t - \epsilon^s_t) +  \sqrt{1 - \alpha_{t-1}} \left( \epsilon + \frac{h^{d}_{t}}{\ell(h^{d})} \right)$
\end{algorithmic}
\end{algorithm}

As the structural variations progressively stabilize, the edited regions of the image become increasingly consistent and precisely localized.
Structural semantics captured by $M^*$ attain high fidelity, while the texture contained in the cross-step entropy acquire enhanced regional discriminative value.
Therefore, we define a fine-grained temporal coopetition noise, as detailed below:
\begin{equation} \label{eq:attn_dds}
    \epsilon^{dir}_{t} = \sqrt{\frac{1-\alpha_t}{\alpha_t}} (\epsilon^e_t-\epsilon^s_t+\frac{h^{d}_{t}- \text{mean}(h^{d}_{t})}{\text{std}(h^{d}_{t})}).
\end{equation}
Here, $(\epsilon^e_t - \epsilon^s_t)$ denotes the noise prediction discrepancy between the editing and reconstruction branches, while $\text{mean}(*)$ and $\text{std}(*)$ refer to the mean and standard deviation of the cross-step entropy divergence.
This normalization aligns $h^{d}_{t}$ with the distribution of predicted noise, enabling effective temporal modulation of the latent states and preserving trajectory consistency.
Subsequently, single step loss $L$ of latent refinement is conducted based on the direction, formulated as:
\begin{gather}
	L = \epsilon_t^{dir} \cdot z^{e}_t\cdot M^*, \\
    z_t^e = z_t^e - \nabla_{z_t^e}L + h^{d}_{t}.
\end{gather}
Here, the gradient update $\nabla_{z_t^e} L$ is applied exclusively to the editable regions indicated by $M^*$, thus preserving background integrity.
In addition, $h^{d}_{t}$is directly incorporated as a regularization term to further enhance the texture details.

The joint process of structure refinement and texture reinforcement ensures coherent semantic progression, yielding the temporally adjusted latent state \(z^e_{t-1}\) for stable editing, as formulated below:
\begin{align} 
z_{t-1} &= \sqrt{\alpha_{t-1}}z_0 + \sqrt{\alpha_{t-1}}(z^e_t-z^s_t)M^*_{t} \\ \nonumber
& \quad + \sqrt{(1-\alpha_t)\alpha_{t-1}/\alpha_t}(\epsilon^e_t-\epsilon^s_t) \\ \nonumber
& \quad + \sqrt{1-\alpha_{t-1}}(\epsilon+\frac{h^{d}_{t}}{\ell(h^{d})}).
\end{align}

By jointly refining structural and textural consistency across timesteps, CoEdit facilitates the minimization of the objective in Eq.~\ref{eq:coop_objective} over the entire sampling horizon \( \mathcal{T} \).

\begin{figure*}[tb]
    \centering
    \includegraphics[width=1\linewidth]{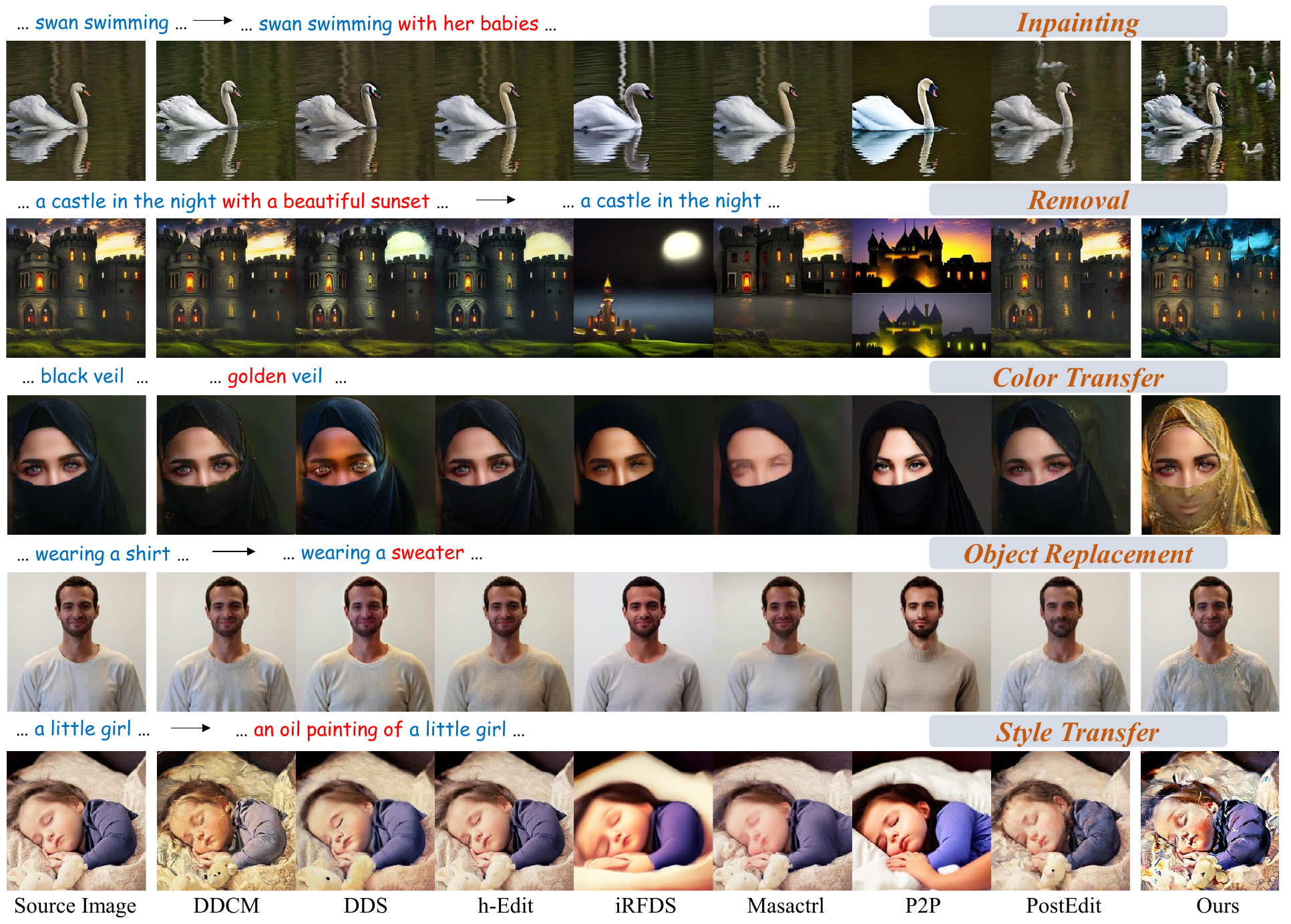}
    \caption{Comparative visualization of various zero-shot image editing methods.}
    \label{fig:pie_demos}
\end{figure*}

\begin{table*}[t]
\caption{Comprehensive quantitative comparison on PIEBench. ${}^\dagger$ indicates adapted results for fair comparison. Best values are \textbf{bold}, second best are \underline{underlined}.}
\label{tab:full_comparison_pie}
\centering
\renewcommand{\arraystretch}{1.2}
\resizebox{\textwidth}{!}{
\begin{tabular}{lcccccccccccc}
\hline
\multirow{2}{*}{{Methods}} & 
\multicolumn{6}{c}{{Default Setting}} & 
\multicolumn{3}{c}{{Editing Setting}} & 
\multicolumn{3}{c}{{Reconstruction Setting}}\\
\cmidrule(lr){2-7} \cmidrule(lr){8-10} \cmidrule(lr){11-13}
& $\text{FCES}\uparrow$ & $\text{CS}_i\uparrow$ & $\text{CS}_r\uparrow$ & $\text{PSNR}\uparrow$ & $\text{LPIPS}\downarrow$ & $\text{SSIM}\uparrow$ & 
$\text{FCES}\uparrow$ & $\text{CS}_i\uparrow$ & $\text{CS}_r\uparrow$ & $\text{PSNR}\uparrow$ & $\text{LPIPS}\downarrow$ & $\text{SSIM}\uparrow$ \\
\cmidrule(lr){1-7} \cmidrule(lr){8-10} \cmidrule(lr){11-13}
P2P & 22.78 & \underline{25.27} & 22.56 & 17.87 & 0.2089 & 0.7164 & 21.95 & 25.43 & 22.63 & 19.49 & 0.1763 & 0.7466 \\
PnP & 22.78 & 25.57 & \underline{22.57} & 22.32 & 0.1127 & 0.7958 & 25.23 & 25.99 & 22.95 & 24.21 & 0.0882 & 0.8193 \\
Pix2PixZero & 24.31& 22.92& 20.63& 20.42& 0.1693& 0.7538 & 22.95& 22.29& 20.40& 22.52& 0.1302& 0.7880\\
NTI & 29.15 & 24.93 & 21.91 & 27.21 & 0.0588 & 0.8483 & 28.44 & 25.10 & 22.11 & 28.19 & 0.0469 & 0.8599 \\
NPI & 28.61 & 24.78 & 21.89 & 26.30 & 0.0680 & 0.8406 & 28.41 & 25.02 & 22.08 & 26.76 & 0.0637 & 0.8445 \\
DI & 29.28 & 24.99 & 22.09 & 27.27 & \underline{0.0540} & \underline{0.8532} & \underline{28.91} & 25.35 & 22.40 & 28.11 & 0.0470 & \underline{0.8608} \\
MasaCtrl & 26.06 & 24.26 & 21.20 & 22.19 & 0.1055 & 0.8032 & 25.43 & 24.70 & 21.55 & 23.70 & 0.0892 & 0.8188 \\
DDCM & \underline{29.60} & 24.84 & 22.05 & \underline{28.06} & 0.0566 & 0.8516 & \textbf{28.95}& 25.47& 22.74& \underline{28.77} & 0.0503 & 0.8595 \\
PostEdit$^\dagger$ & 27.51 & 25.12 & 22.15 & 25.41 & 0.0932 & 0.7983 & 26.70 & 25.64 & 22.60 & 26.47 & 0.0801 & 0.8125 \\
iRFDS$^\dagger$ & 24.16 & \textbf{25.97} & \textbf{22.76} & 19.66 & 0.1691 & 0.7437 & 23.46 & 26.02 & \underline{23.18} & 21.05 & 0.1453 & 0.7658 \\
h-Edit$^\dagger$ & 29.25 & 25.45 & 22.41 & 26.62 & 0.0545 & 0.8470 & 27.47 & \underline{26.09} & 23.17 & 27.90 & \underline{0.0446} & 0.8581 \\
\hline
\textbf{CoEdit} & \textbf{29.81} & 24.60 & 21.68 & \textbf{28.28} & \textbf{0.0513} & \textbf{0.8574} & 27.43 & \textbf{26.13} & \textbf{23.24} & \textbf{29.92} & \textbf{0.0421} & \textbf{0.8700} \\
\hline
\end{tabular}
}
\end{table*}
\renewcommand{\arraystretch}{1.0}

\section{Experiments}
In this section, we perform a comprehensive comparison with state-of-the-art training-free image editing methods, followed by a series of ablation studies designed to validate the contribution of each major component.

\subsection{Datasets and Settings}
\paragraph{Benchmark}
To ensure fair and representative evaluation, we follow mainstream SOTA work~\cite{xu2023infedit, ju2024pnp} and adopt PIEBench as the primary benchmark. 
PIEBench is a large-scale benchmark that comprises over 700 instances across 9 distinct categories, each entry is annotated with a source prompt, target prompt, natural language instruction, and a pixel-level binary mask of the modified region.
The availability of pixel-level annotation masks, along with the dataset's high diversity, makes it particularly suitable for quantitatively evaluating spatial precision and semantic alignment in image editing tasks.
These annotation masks are used only for post-hoc evaluation, such as region-aware metrics and mask-quality analysis; they are never provided to CoEdit during inference. 
In addition, we present experimental results on the PIEBench++ \cite{piepp} benchmark.

\paragraph{Experimental settings}
In text-guided image editing, a fundamental trade-off exists between the semantic accuracy of modifications and the fidelity of structural preservation.
We adopt a protocol in line with prior works such as h-Edit \cite{hedit} and PostEdit \cite{PostEdit}, establishing three distinct operational regimes. 
These regimes are exclusively modulated by the classifier-free guidance scale, corresponding to: (1) a default setting for balanced performance, (2) an editing-centric setting to probe the upper bound of semantic manipulation, and (3) a reconstruction-centric setting to ascertain the upper bound of fidelity.
For the denoising process, we employ the LCM-SD1.5 as denoising model. 
In alignment with the majority of attention-based editing methods, we selectively manipulate the cross-attention layers in the deeper stages.
Within spatial coopetition, we employ L2 norm as coopetition norm function $\ell(*)$. 
In temporal coopetition, $N_o$ is set to 50.

\paragraph{Evaluation metrics}
To comprehensively evaluate the performance of image editing algorithms, we adopt a multi-faceted evaluation framework that includes a variety of metrics, such as PSNR, LPIPS, SSIM, CLIP-based Similarity of image ($\text{CS}_i$) and CLIP-based Similarity of edited region ($\text{CS}_r$) of the generated images. 

\begin{figure}[t]
    \centering
    \includegraphics[width=1.0\linewidth]{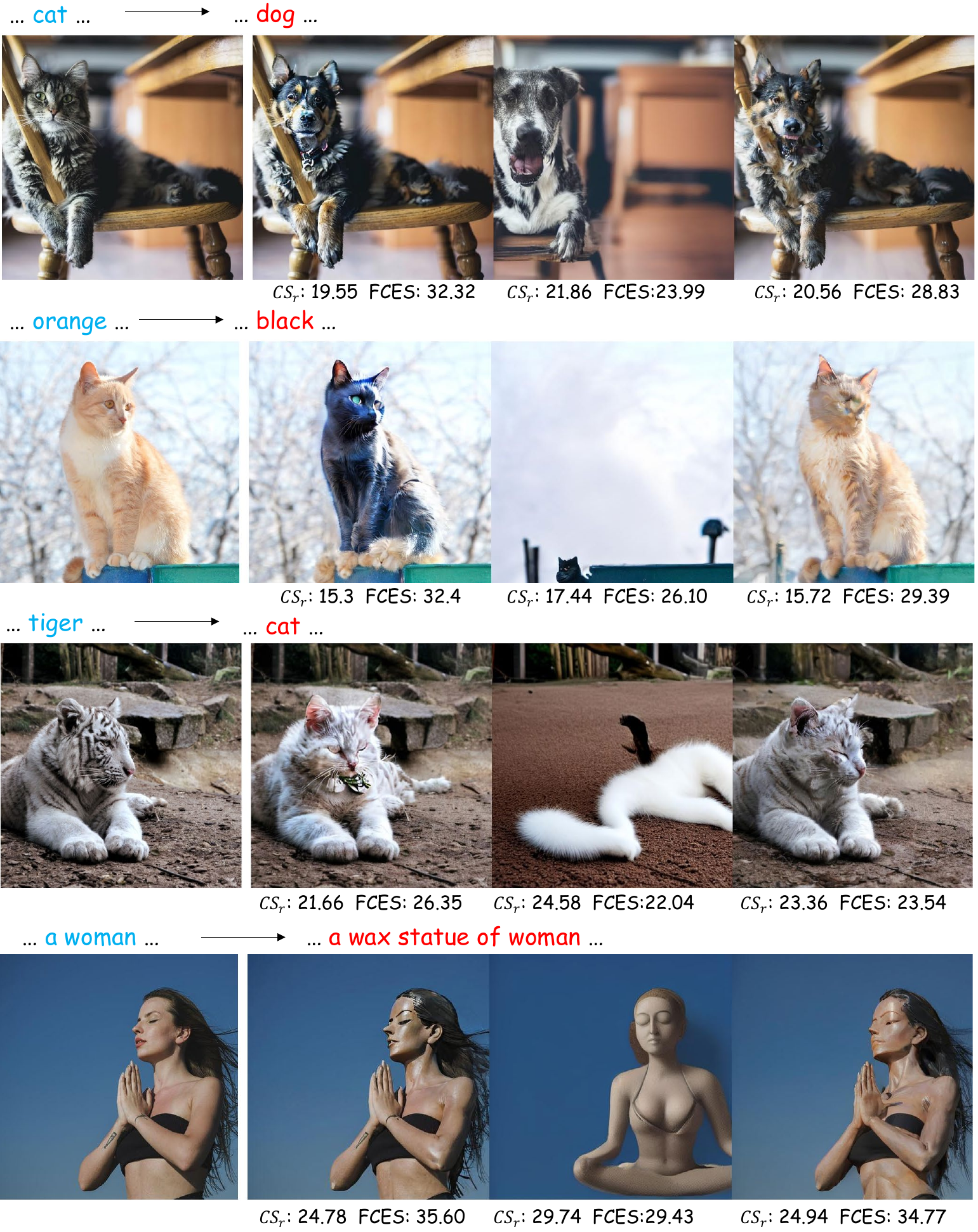}
    \caption{Visualization of failure examples with higher $CS_r$.}
    \label{fig:wrong_cs}
\end{figure}

However, these metrics alone are insufficient to provide a holistic assessment, as they fail to simultaneously capture fidelity in the reconstructed regions and diversity in the edited regions, as shown in Fig. \ref{fig:wrong_cs}.
To address this trade-off and account for the relative proportions of edited and unedited regions within an image, we introduce a more comprehensive metric, termed the Fidelity-Constrained Editing Score (FCES):
\begin{equation}
\label{eq:fces}
\begin{split}
\text{FCES} = & \,\, \lambda \big( w_e \cdot \text{CS}_{r} + w_e \cdot \text{CS}_{i} \\
& + w_s \cdot \text{PSNR} / S_p + w_s \cdot \text{SSIM} \big),
\end{split}
\end{equation}

where $\lambda$ denotes the scaling factor, $w_e$ and $w_s$ denote the spatial proportions of the edited and reconstructed regions within the entire image.
As with other region-aware evaluation metrics, $w_e$ and $w_s$ are computed from benchmark annotations only at evaluation time and are not used by CoEdit during inference.
We select two metrics to evaluate each aspect of editing performance: CS$_r$(Editing Region Clip Score) and CS$_i$(Entire Image Clip Score) for semantic alignment and instruction consistency in the edited regions, and PSNR and SSIM for fidelity in the reconstructed regions.
PSNR is normalized by a scaling factor $S_p$ (40dB), as this value is widely considered to correspond to perceptually lossless image quality.

\subsection{Performance Comparison}

\paragraph{Evaluation methods}
We conduct comprehensive comparative experiments against a diverse array of state-of-the-art zero-shot image editing methodologies, including P2P \cite{ptop}, PnP \cite{pnp},Pix2PixZero \cite{pix2pixzero}, NTI \cite{NTI}, NPI \cite{NPI}, DI, MasaCtrl \cite{masactrl}, and DDCM \cite{xu2023infedit}, along with recent advancements, for instance, PostEdit \cite{PostEdit}, iRFDS \cite{iRFDS}, and h-Edit \cite{hedit}.
To ensure the fairness of experiments, we adopt the same experimental settings (random seed is 0), and conduct all experiments on 8 NVIDIA A6000 GPUs. The experimental of this work was supported by the Centre for Large AI Models (CLAIM) of the Hong Kong Polytechnic University.

\paragraph{Comparison on PIEBench}
Table~\ref{tab:full_comparison_pie} presents a comprehensive quantitative comparison on PIEBench, validating the superior performance of CoEdit in training-free image editing. It highlights the effectiveness of our spatial and temporal coopetition strategy in maintaining background integrity during modifications.
Even under high CFG strength in editing setting, CoEdit retains strong editing performance, consistently outperforming advanced baselines. Under the stringent reconstruction setting (low CFG strength), it demonstrates comprehensive dominance across all metrics. These findings empirically confirm that transforming attention control into a coopetitive strategy yields state-of-the-art performance across all operational scenarios and key evaluation criteria.

\paragraph{Comparison on PIEBench++}
To ensure a fair comparison on PIE-Bench++, where parameters of various methods are not directly comparable or optimal, we evaluate all methods under distinct editing and reconstruction settings. 
As presented in Table~\ref{tab:pie_pp_comparison}, this dual evaluation highlights CoEdit's superior balance. Under the editing-centric configuration, CoEdit achieves the highest CLIP similarity scores. Concurrently, in the reconstruction setting, it attaining the highest PSNR of 30.69, which underscores CoEdit's precise control over the editing-fidelity trade-off.

\paragraph{User study}
To further assess the human alignment of our proposed framework, we conducted a comprehensive user study, as illustrated in Table \ref{tab:user_study}. 
Edited images are assessed across three aspects: Image Fidelity, Editing Quality, and Overall Score, capturing both perceptual realism and semantic alignment. 
Each aspect is scored on a five‑point Likert scale, where a score of 1 denotes the lowest quality and 5 represents the highest.
The evaluation comprises 7,000 annotations collected from 10 participants. 
Results confirm that CoEdit’s spatial–temporal coopetition strategy not only enhances objective metrics but also substantially improves subjective user experience.

\begin{table}[t]
\caption{Quantitative comparison of training-free image editing methods on PIEBench++.}
\label{tab:pie_pp_comparison}
\centering
\renewcommand{\arraystretch}{1.2}
\resizebox{0.95\linewidth}{!}{
\begin{tabular}{lcccccc}
\hline
\multirow{2}{*}{{Methods}} &
\multicolumn{3}{c}{{Editing Setting}} &
\multicolumn{3}{c}{{Reconstruction Setting}} \\
\cmidrule(lr){2-4} \cmidrule(lr){5-7}
 & $\text{FCES}\uparrow$ & $\text{CS}_i\uparrow$ & $\text{CS}_r\uparrow$ & $\text{PSNR}\uparrow$ & $\text{LPIPS}\downarrow$ & $\text{SSIM}\uparrow$ \\
\hline
P2P                      & 21.15 & 24.18 & 23.77 & 20.67 & 0.1426 & 0.7917 \\
PnP                      & 23.69 & \underline{26.18} & \underline{25.65} & 25.19 & 0.0713 & 0.8529 \\
Pix2Pix-Zero             & 20.73 & 21.73& 21.44& 22.86& 0.1195& 0.8148\\
NTI                      & 25.39 & 24.19& 23.64& \underline{29.65}& \textbf{0.0330}& \textbf{0.8920}\\
NPI                      & 25.44 & 24.21& 23.64& 27.14& 0.0550& 0.8701\\
DI                       & \textbf{25.73} & 24.32& 23.78& 29.44& \underline{0.0341}& \underline{0.8905}\\
MasaCtrl                 & 23.91 & 24.11& 23.56& 22.35& 0.0978& 0.8291\\
$\text{PostEdit}$        & \underline{25.51} & 25.35& 24.78& 27.27& 0.0669& 0.8421\\
$\text{iRFDS}$           & 23.00 & 26.08 & 25.45 & 22.39 & 0.1150 & 0.8072 \\
$\text{h-Edit}$          & 25.01 & 25.65 & 25.09 & 28.41& 0.0378& 0.8829 \\
\hline
\textbf{CoEdit}          & 25.11 & \textbf{26.49} & \textbf{25.85} & \textbf{30.69} & 0.0387& 0.8872 \\
\hline
\end{tabular}
}
\end{table}
\renewcommand{\arraystretch}{1.0}

\subsection{Ablation Studies of CoEdit}
\paragraph{Ablation study of coopetitive components}
We conduct an ablation study to assess the contributions of the spatial and temporal coopetition strategies within CoEdit. As shown in Table~\ref{tab:ablation_module}, removing the spatial coopetition module, implemented via Dual-Entropy Attention Manipulation, results in reduced performance across both editing and reconstruction metrics. Similarly, eliminating the temporal coopetition module, realized through Temporal Entropic Latent Refinement, leads to noticeable degradation in fidelity and consistency. These results demonstrate that both components are essential for achieving balanced and high-quality editing outcomes.

\begin{table}[htb]
\centering
\renewcommand{\arraystretch}{1.2}
\caption{Ablation of spatial and temporal coopetition.}
\resizebox{1.0\columnwidth}{!}{
\begin{tabular}{lcccccc}
\toprule
Methods & $\text{FCES}_{\%}\uparrow$ & $\text{CS}_{i,\%}\uparrow$ & $\text{CS}_{r,\%}\uparrow$ & $\text{PSNR}_{\%}\uparrow$ & $\text{LPIPS}\downarrow$ & $\text{SSIM}\uparrow$ \\ 
\midrule
\textbf{w/o} spatial coopetition & 27.87 & \textbf{25.82} & \textbf{22.85} & 24.77 & 0.0777 & 0.8229 \\
\textbf{w/o} temporal coopetition & 29.09 & 25.41 & 22.49 & 26.87 & 0.0621 & 0.8441 \\
\rowcolor{gray!25} CoEdit(default) & \textbf{29.81} & 24.60 & 21.68 & \textbf{28.28} & \textbf{0.0513} & \textbf{0.8574} \\
\bottomrule
\end{tabular}
}
\label{tab:ablation_module}
\renewcommand{\arraystretch}{1.0}
\end{table}

\begin{table}[htb]
\centering
\renewcommand{\arraystretch}{1.2}
\caption{Ablation study of coopetition hyperparameters.}
\resizebox{1.0\columnwidth}{!}{
\begin{tabular}{lccccccc}
\toprule
Type & Methods & $\text{FCES}_{\%}\uparrow$ & $\text{CS}_{i,\%}\uparrow$ & $\text{CS}_{r,\%}\uparrow$ & $\text{PSNR}_{\%}\uparrow$ & $\text{LPIPS}\downarrow$ & $\text{SSIM}\uparrow$ \\
\midrule
\multirow{4}{*}{Spatial} & Frobenius & 29.72 & 24.71 & 21.83 & 28.17 & 0.0531 & 0.8558 \\
& $L_1$ & 30.62 & 24.06 & 21.10 & 29.80 & 0.0431 & 0.8688 \\
& $L_\infty$ & 27.69 & 25.44 & 22.44 & 24.39 & 0.0813 & 0.8184 \\
\rowcolor{gray!25} & \begin{tabular}{@{}c@{}}$L_2$ \\ $N_o=50$\end{tabular} & 29.81 & 24.60 & 21.68 & 28.28 & 0.0513 & 0.8574 \\
\midrule
\multirow{2}{*}{Temporal} & $N_o=100$ & 29.73 & 24.69 & 21.85 & 28.17 & 0.0537 & 0.8554 \\
& $N_o=150$ & 29.30 & 24.97 & 22.05 & 27.43 & 0.0588 & 0.8494 \\
\bottomrule
\end{tabular}
}
\label{tab:ablation_co_parameters}
\renewcommand{\arraystretch}{1.0}
\end{table}

\begin{figure}[t]
    \centering
    \includegraphics[width=0.85\linewidth]{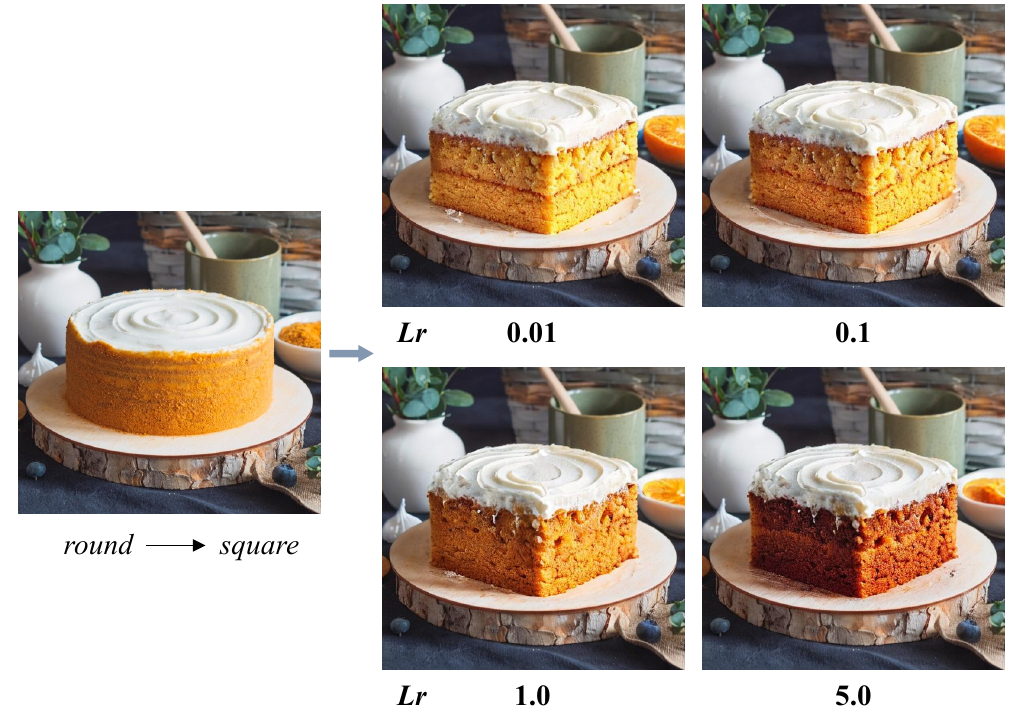}
    \caption{Visualization of different learning rate.}
    \label{fig:lr_visual}
\end{figure}

\begin{table}[htb]
\caption{User study results reporting average scores for Fidelity, Editing Quality, and Overall Score.}
\label{tab:user_study}
\renewcommand{\arraystretch}{1.2}
\setlength{\tabcolsep}{10pt}
\centering
\resizebox{0.9\columnwidth}{!}{
\begin{tabular}{lccccc}
\hline
{Metric} & CoEdit & h-Edit & PostEdit & DDCM & iRFDS \\
\hline
Fidelity        & \textbf{3.93} & 3.72 & 3.56 & 3.71 & 2.46 \\
Editing         & \textbf{3.13} & 3.05 & 2.48 & 2.84 & 2.44 \\
Overall         & \textbf{3.32} & 3.12 & 2.58 & 2.94 & 2.23 \\
\hline
\end{tabular}
}
\end{table}
\renewcommand{\arraystretch}{1.0} 

\begin{figure*}[t]
    \centering
    \includegraphics[width=1.0\linewidth]{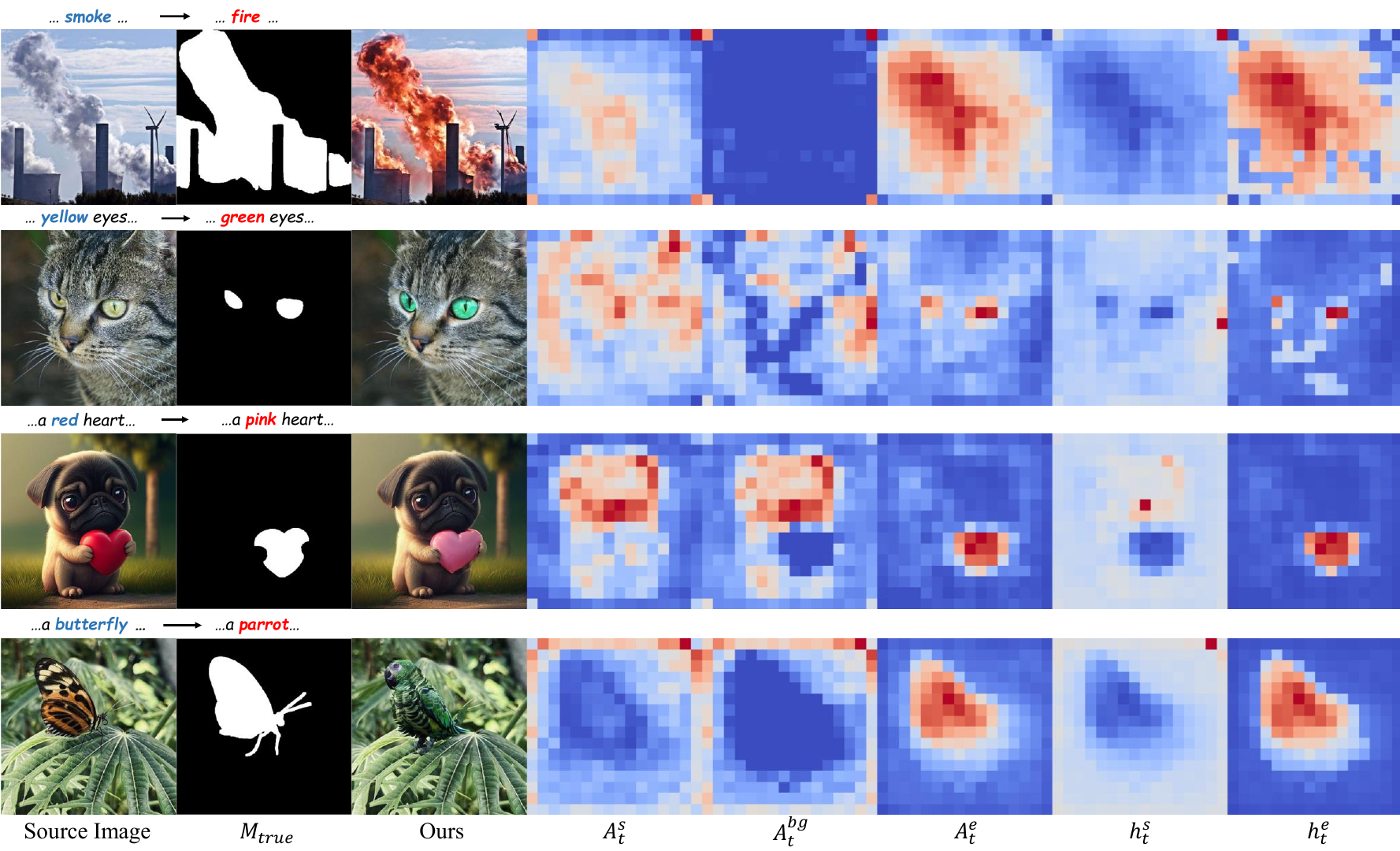}
    \caption{Illustration of original attention maps and dual entropy, demonstrating the spatial coopetitive strategy.}
    \label{fig:de_demo}
\end{figure*}

\paragraph{Ablation study of spatial coopetition}
To assess the effectiveness of our spatial coopetition strategy in disentangling attention conflicts, Fig.~\ref{fig:de_demo} presents a comparison of spatial coopetition, between dual entropy and raw attention scores. While the raw attention maps \( A^s_t \) and \( A^e_t \) exhibit overlapping and unstable patterns due to conflicting semantic claims, the directional entropy maps \( h^s_t \) and \( h^e_t \) demonstrate complementary spatial emphasis. 
This mutual guidance clarifies the separation of editable and preservable regions and improves both structural coherence and textural fidelity, which collectively reflect the cooperative behavior promoted by our method.

\paragraph{Ablation study of coopetition hyperparameters}
Table~\ref{tab:ablation_co_parameters} further validates the hyperparameter choices for our spatial and temporal coopetition. 
For the spatial norm, the data clearly replicates the inherent conflict of competitive strategies: the $L_1$ norm excels at reconstruction by sacrificing edit alignment. 
Conversely, the $L_\infty$ norm maximizes $\text{CS}$ scores but causes a severe collapse in reconstruction quality (PSNR $24.39\%$). 
Our default {$L_2$ norm} achieves the optimal coopetitive balance, maintaining high fidelity across all reconstruction metrics and attaining a near-optimal FCES score.
For the temporal hyperparameter $N_o$, setting it to $100$ or $150$ both result in performance degradation compared to our default configuration, particularly the $N_o=150$ setting, which shows a significant drop in FCES and PSNR. This confirms the efficacy of our default temporal parameterization.
Moreover, to provide a clear illustration of this temporal mechanism, the dynamic learning rate schedule governing our temporal coopetition is visualized in Fig. \ref{fig:lr_visual}

\begin{figure}[t]
    \centering
    \includegraphics[width=1.0\linewidth]{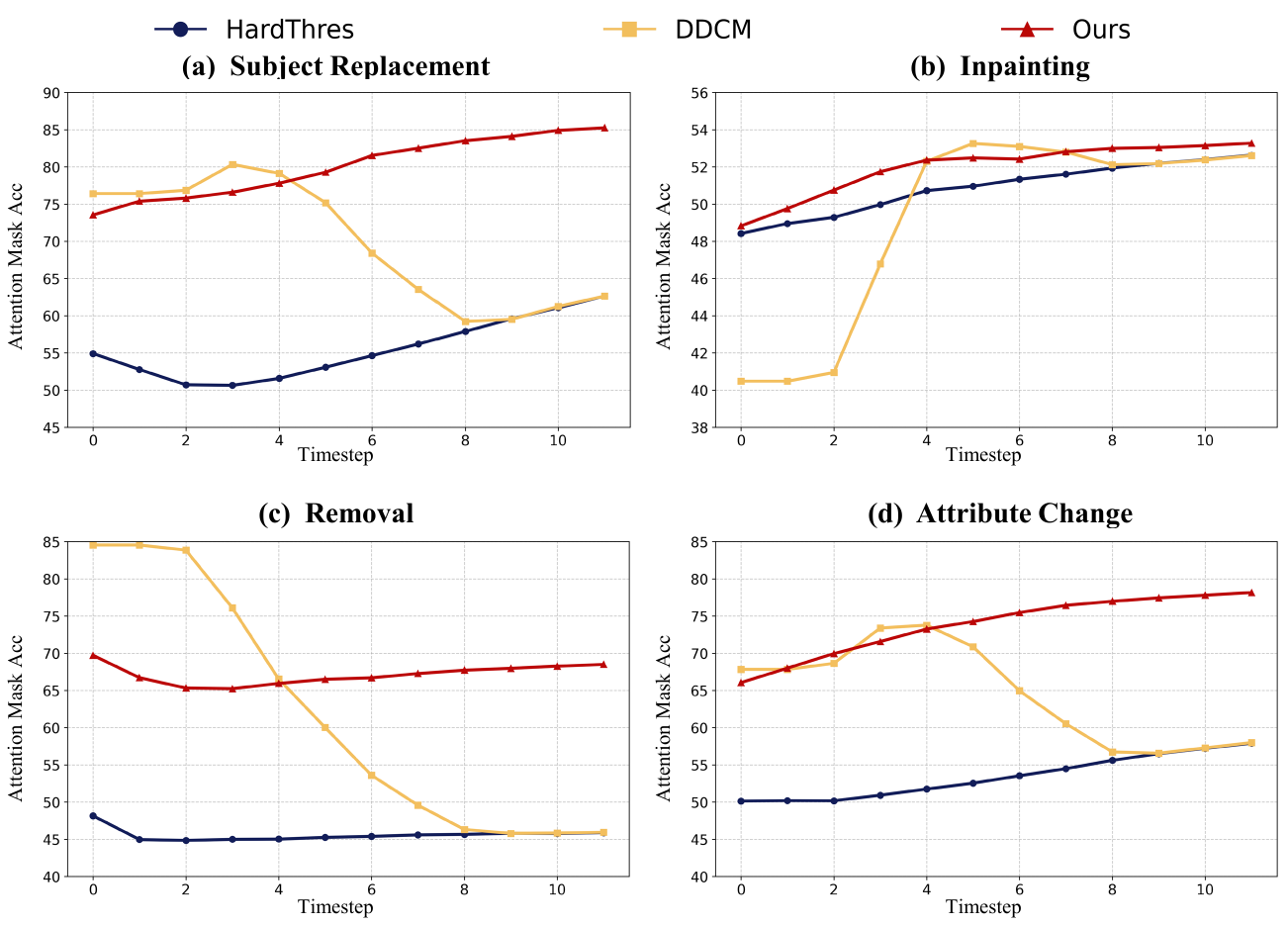}
    \caption{Post-hoc comparison of attention mask accuracy throughout denoising steps among our method, DDCM, and hard threshold approaches using benchmark annotations only for evaluation. The temporal coopetitive strategy effectively enhances the discriminative capacity between editing and reconstruction regions.}
    \label{fig:attn_mask}
\end{figure}

\paragraph{Ablation study of temporal coopetition}
To demonstrate the effectiveness of our temporal cooperation strategy, we conduct a post-hoc analysis on PIEBench by comparing the predicted editing mask $M^*_t$ with the benchmark ground-truth mask across all sampling steps.
These annotation masks are used strictly for evaluation and are not involved in the inference process of CoEdit.
For baseline comparison, we used the hard threshold method (set at 0.3) and DDCM. 
While DDCM utilizes a classical dynamic threshold that reverts to a fixed threshold truncation strategy in later sampling stages, this causes its performance to closely resemble that of a hard threshold baseline.
Fig. \ref{fig:attn_mask}(a)-(d) illustrate four editing scenarios: object replacement, inpainting, removal, and attribute modification, respectively.
During  early stages of sampling, the latent representation is close to random noise, resulting in less reliable attention masks.
Our method, CoEdit, progressively improves segmentation accuracy throughout the sampling process.
\subsection{Visualization Demos}
The visualization examples in Fig.~\ref{fig:pie_demos} show a wide range of  editing tasks, including inpainting, object removal, color transfer, style transformation, and object replacement. The results across all methods highlight the ability of our framework to preserve structural coherence while achieving semantically accurate and visually diverse edits across various prompt types.

\section{Conclusion}
In this work, we present a novel zero-shot framework, namely CoEdit, that redefines attention control as a coopetitive process rather than a purely competitive one. 
It boosts editing and reconstruction branches to dynamically negotiate semantic influence through entropy-guided cooperation. 
Spatially, Dual-Entropy Attention Manipulation disentangles attention maps by quantifying directional entropy, allowing precise partitioning of editable and reconstructable regions.
Temporally, Entropic Latent Refinement ensures consistent semantic alignment across timesteps.
Furthermore, we introduce FCES to jointly reflects edit diversity and reconstruction quality.
Extensive experiments on standard benchmarks demonstrate that CoEdit achieves SOTA in both editing accuracy and structural fidelity.

\bibliographystyle{IEEEtran} 
\bibliography{reference}

\vfill

\end{document}